\documentclass[letterpaper]{article} 
\usepackage{aaai25}  
\usepackage{times}  
\usepackage{helvet}  
\usepackage{courier}  
\usepackage[hyphens]{url}  
\usepackage{graphicx} 
\usepackage{natbib}  
\usepackage{caption} 
\usepackage{algorithm}
\usepackage{algorithmic}
\usepackage{graphicx}
\usepackage{multirow}
\usepackage{booktabs}
\usepackage{cite}
\usepackage{color}
\usepackage[dvipsnames]{xcolor} 
\usepackage{blindtext}
\usepackage{card}
\usepackage{newfloat}
\usepackage{listings}
\DeclareCaptionStyle{ruled}{labelfont=normalfont,labelsep=colon,strut=off} 
\floatstyle{ruled}
\newfloat{listing}{tb}{lst}{}
\floatname{listing}{Listing}
\title{Y-Mol: A Multiscale Biomedical Knowledge-Guided Large Language Model for Drug Development}
\author{
    Tengfei Ma\textsuperscript{\rm 1},
    Xuan Lin\textsuperscript{\rm 2},
    Tianle Li\textsuperscript{\rm 3}, 
    Chaoyi Li\textsuperscript{\rm 1}, 
    Long Chen\textsuperscript{\rm 2}, 
    Peng Zhou\textsuperscript{\rm 1}, 
    Xibao Cai\textsuperscript{\rm 1}, 
    Xinyu Yang\textsuperscript{\rm 1}, 
    Daojian Zeng\textsuperscript{\rm 3}, 
    Dongsheng Cao\textsuperscript{\rm 4}, 
    Xiangxiang Zeng\textsuperscript{\rm 1}
}
\affiliations{
    \textsuperscript{\rm 1}College of Computer Science and Electronic Engineering, Hunan University, China \\
    \textsuperscript{\rm 2}Key Laboratory of Intelligent Computing and Information Processing
Ministry of Education, Xiangtan University, China \\
    \textsuperscript{\rm 3}Hunan Provincial Key Laboratory of Intelligent Computing and Language Information Processing, Hunan Normal University, China \\
    \textsuperscript{\rm 4}Xiangya School of Pharmaceutical Sciences, Central South University, China
}

\usepackage{bibentry}

\begin{document}

\maketitle

\begin{abstract}

Large Language Models (LLMs) have recently demonstrated remarkable performance in general tasks across various fields. However, their effectiveness within specific domains such as drug development remains challenges. To solve these challenges,  we introduce \textbf{Y-Mol}, forming a well-established LLM paradigm for the flow of drug development. Y-Mol is a multiscale biomedical knowledge-guided LLM designed to accomplish tasks across lead compound discovery, pre-clinic, and clinic prediction. By integrating millions of multiscale biomedical knowledge and using LLaMA2 as the base LLM, Y-Mol augments the reasoning capability in the biomedical domain by learning from a corpus of publications, knowledge graphs, and expert-designed synthetic data. The capability is further enriched with three types of drug-oriented instructions: description-based prompts from processed publications, semantic-based prompts for extracting associations from knowledge graphs, and template-based prompts for understanding expert knowledge from biomedical tools. Besides, Y-Mol offers a set of LLM paradigms that can autonomously execute the downstream tasks across the entire process of drug development, including virtual screening, drug design, pharmacological properties prediction, and drug-related interaction prediction. Our extensive evaluations of various biomedical sources demonstrate that Y-Mol significantly outperforms general-purpose LLMs in discovering lead compounds, predicting molecular properties, and identifying drug interaction events. The source code is available at https://anonymous.4open.science/r/Y-Mol.
\end{abstract}

\section{Introduction}

Large Language Models (LLMs) including GPT-4~\cite{openai2023gpt}, ChatGLM~\cite{zeng2022glm}, and LLaMA~\cite{touvron2023llama} have achieved great success in various applications~\cite{fang2023mol,lyu2023llm} due to their powerful reasoning capability. These models boasting billions of parameters, are meticulously trained on large text corpora and excel at generating human-interactable text and easy-to-understand contexts. While LLMs have shown promising performance in general tasks, they are limited in the capability of domain applications~\cite{zhang2023biomedgpt,zhou2024instruction,10.1093/jamia/ocae037}.
In order to make LLM applicable to specific domains and tasks, a large number of researchers have designed task-relevant instructions to efficiently fine-tune the parameters of the model~\cite{ouyang2022training,sanh2021multitask}. This process entails training models using specialized instruction datasets. By doing so, the models acquire task-specific knowledge and patterns, that improve their performance in targeted domains~\cite{fang2023mol}. Various instruction datasets have been developed in general domains~\cite{taori2023stanford,tinn2023fine,wang2022self}. However, a major barrier to harnessing LLMs in the drug development domain is the lack of a dedicated dataset~\cite{fang2023mol}. 
The gap is primarily raised from \textbf{three challenges:} \textbf{(i)} acquiring drug-related data is significantly costly and the general knowledge of drug development spans a broad spectrum including computational chemistry, structural biology, and bioinformatics. \textbf{(ii)} The intrinsic interaction data between biomedical entities such as the drug-perturbed gene expression and protein binding activity need fine-grained domain knowledge, given the inherent complexity of annotating. \textbf{(iii)} Different from the well-established frameworks in natural language processing, there is no standard paradigm for drug development. Different scenarios apply various representations for drug-related information, amplifying the challenges of crafting a dataset based on LLM across the domain of drug development.

\begin{figure}
\centering %
\includegraphics[width=1\columnwidth]{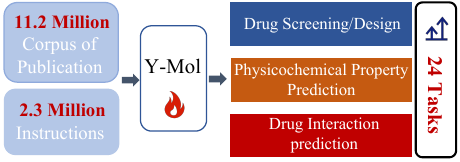} %
\caption{Y-Mol provides large-scale corpus and instructions for drug development 
across 24 tasks.}
\label{fig:intro}
\end{figure}

To overcome these limitations, we propose a multiscale biomedical knowledge-guided LLM to enhance the potential of drug development, called Y-Mol. Y-Mol is implemented as an autoregressive sequence-to-sequence model finetuned on LLaMA2~\cite{touvron2023llama} over different text corpus and instructions derived from various biomedical knowledge (Figure~\ref{fig:modules}). Specifically, we construct a large-scale biomedical text corpus from PubMed publications across various domains (e.g., bioinformatics) related to drug development, which alleviates the significant data acquisition and annotation costs (respond to challenge \textbf{(i)}). We build biomedical interaction instructions from a large-scale knowledge graph using well-designed prompts to model the rich-depth interactions between biomedical entities and the drug-perturbed expression data (for challenge \textbf{(ii)}). To unify different drug development applications, we distill expert knowledge extracted from existing small models into Y-Mol, enhancing the consistencies of representations for drug-related information. To achieve this, we design prompt templates and construct a set of instructions based on the synthesized data from small models (e.g., the small prediction models for ADMETs~\cite{xiong2021admetlab,zeng2022accurate} and drug repurposing~\cite{zeng2019deepdr,zeng2020repurpose}), which provides a reliable knowledge spectrum to promote the capabilities of drug development (address challenge \textbf{(iii)}). Based on the constructed instructions, we consider LLaMA2 as the base LLM and fine-tune it to enhance its capabilities in drug development. To evaluate the effectiveness of Y-Mol on drug development, we intuitively design various tasks across lead compound discovery, pre-clinic, and clinic predictions. The results highlight the value of constructed drug instructions and demonstrate the ability of Y-Mol to enhance the versatility and understanding of the LLM paradigm for drug development.


Our key contributions can be concluded as follows:
\begin{itemize}
    \item To the best of our knowledge, \textbf{Y-Mol} is the first to construct an LLM paradigm for drug development.
    \item To alleviate the high costs of data acquiring for LLM-enhanced drug development, \textbf{Y-Mol} leverages multiscale biomedical knowledge to construct an informative instruction dataset and unifies the overflow of drug development, by collecting the corpus of publications, domain knowledge graph, and synthesized data from small models across lead compound discovery, pre-clinic, and clinic predictions.

    \item Extensive experiments on drug development show the performance of building instruction datasets, demonstrating \textbf{Y-Mol} have powerful capabilities to enhance the understanding and universality of various tasks.
\end{itemize}
\section{Related Work}
\subsubsection{Data-centric Instruction Models.} \textcolor{black}{The major barrier to applying LLM in drug development is the lack of a dedicated dataset. Previous studies have intensively grasped the meta-knowledge within and across different information sources. For example, MolT5 \cite{edwards-etal-2022-translation} is the first to link molecules to natural language by using 33K molecule-description pairs as finetuning and evaluation data. KV-PLM \cite{zeng2022deep} injects SMILES sequences into description-based text by performing masked language modeling on 10K structure-text pairs, and MolXPT \cite{liu-etal-2023-molxpt} further localizes molecule names to obtain a mixed corpus of 30M text, 30M SMILES and 8M wrapped sequences between SMILES and text for pre-training. For structure-text pre-training, MoleculeSTM \cite{liu2023moleculestm} uses 281K structure-text pairs data to encode molecular graphs and text with independent encoders. These datasets are very valuable for pre-training smaller models, but this limits their direct usefulness for LLMs. Recently, Mol-Instruction \cite{fang2023mol} is introduced with a comprehensive instruction dataset designed for three types of biomolecular tasks, which is general in molecular biology. However, their proficiency in drug development, relying on drug-related response data, remains limited.}

\subsubsection{Prompt-centric Instruction Models.} \textcolor{black}{LLMs emerged as useful systems that provide unprecedented opportunities to advance drug development. Some LLM-based studies have been proposed to perform drug design by introducing various prompting paradigms. For example, MolReGPT \cite{li2024empowering} introduces a retrieval-based prompt paradigm for molecule-caption translation. MolecularGPT \cite{liu2024moleculargpt} presents curated molecular instructions spanning over 1000 property prediction tasks. TxT-LLM \cite{chaves2024tx} is proposed to interleave free-text instructions with string representations of molecules spanning various stages of the drug discovery pipeline. The main obstacle in this field is the lack of a unified prompt paradigm for the whole pipeline of drug development, which limits their ability to reason on complex tasks (e.g., drug-target associations) and to generalize to unseen examples (e.g., de novo molecular design). Therefore, it is desired to design a novel paradigm in drug discovery.}
\section{Background}
\subsection{Biomedical Knowledge Graph}
We define a biomedical knowledge graph (KG) as $\mathcal{G}_{kg}=\{(h,r,t)|h,t\in \mathcal{E},r\in\mathcal{R}\}$, where each triple $(h,r,t)$ denotes a relation $r$ between biomedical entities $h$ and $t$. Biomedical KGs contain large-scale interactions between various entities, including structural drug-related gene expression and therapeutic information, which improve the reasoning ability of LLMs for drug development.

\subsection{Knowledge Mining by Models and Tools}
Researchers have developed various computer-assisted models and tools in the flow of drug development. These models simulate the different stages of drug development and they are trained on high-quality data from web-lab (e.g., ADMETlab~\cite{fu2024admetlab}) or physical model (e.g., RDKit~\cite{bento2020open}), which shows generalization ability to current application~\cite{edfeldt2024data}. The data generated from these models can bring us a comprehensive perspective on drug development across different tasks. To fully explore the potential of these methods on drug development, we design a LLM paradigm to model their execution.
\section{Method}
\begin{figure*}
  \centering %
  \includegraphics[width=1\textwidth]{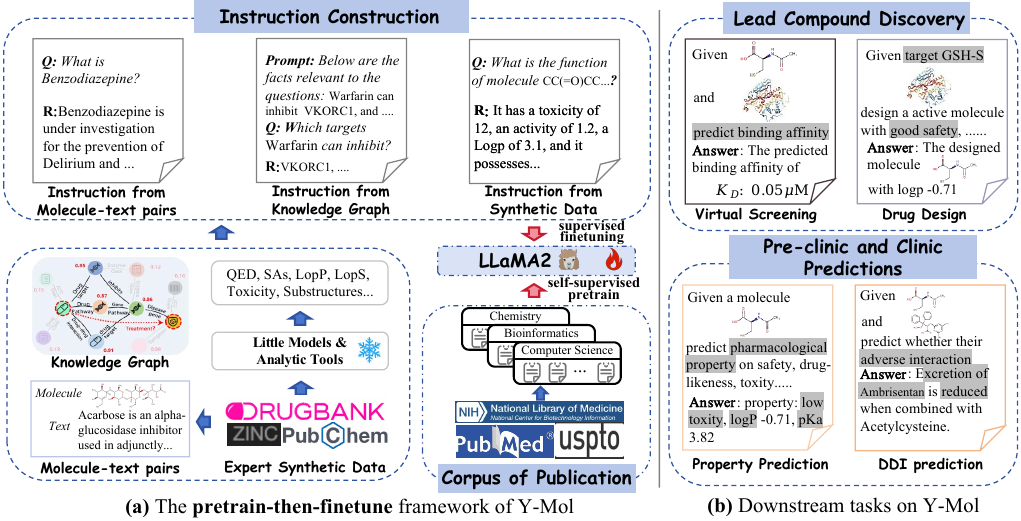} %
  \caption{The architecture of Y-Mol. Y-Mol builds the LLM paradigm for drug development, which comprises two processes: (a) The \textit{pretrain-then-finetune framework} of Y-Mol begins to self-supervised pretrain LLaMA2 based on biomedical publications, then finetune LLaMA2 using constructed instructions; (b) Y-Mol evaluates \textit{downstream tasks} on the finetuned LLaMA2.}
\label{fig:modules}
\end{figure*}

In this section, we start with a brief introduction to our model architecture, followed by the corpus collection from biomedical publications, the instruction construction from molecule-text pairs (i.e., drug databases), the biomedical knowledge graph, and the expert synthetic data. Then, we introduce the strategy of supervised finetuning on the constructed instructions. Finally, we describe our downstream tasks, as well as implementation details.

\subsection{Overview}\label{Sec:overview}
As illustrated in Figure~\ref{fig:modules}, we consider LLaMA2-$7$b as our base LLM to construct a training and reasoning paradigm Y-Mol for drug development. Y-Mol consists of two stages: (a) The \textit{pretrain-then-finetune} framework self-supervised pretrains LLaMA2 on the biomedical corpus of publications, giving Y-Mol a generalized understanding of the background knowledge for drug development. Then the framework supervised finetuning LLaMA2 on drug-related domain knowledge and expert synthetic data across various applications, feeding large-scale drug-related information into Y-Mol to enhance the capabilities of understanding interaction mechanisms for the flow of drug development. (b) Conducting \textit{downstream tasks} for drug development across lead compound discovery, pre-clinic, and clinic predictions based on Y-Mol. To achieve this, Y-Mol provides a paradigm to construct a dataset for pretraining and finetuning LLM following the flow of drug development. The experimental results across lead compound discovery, pre-clinic, and clinic predictions demonstrate the efficiency of Y-Mol.

\subsection{Corpus of Publication}\label{Sec:pretrain}
To fully explore the potential biomedical knowledge from publications, such as the interaction pathway of drugs and the gene expression levels under various compounds, we extract and preprocess over 33 million publications across multiple topics\footnote{The detailed topic refers to Appendix A.1} from online publishers (e.g., PubMed\footnote{https://pubmed.ncbi.nlm.nih.gov/}). As shown in Figure~\ref{fig:dataset}A, we extract their visible abstracts and brief introductions as biomedical texts based on these publications. To align the expressions of drugs and to promote the consistency of biomedical entities, we introduce a named entity recognition (NER) tool~\cite{sung2022bern2} to match entities and use a standard biomedical PubTator~\cite{wei2024pubtator} to modify them. We adopt their SMILES sequence for drug entities to replace entity text, which is beneficial to enhancing the Y-Mol's understanding of molecular language. We consider the processed biomedical texts as the pretraining corpus, which is utilized for self-supervised training following the way of next token prediction. Y-Mol mining potential drug knowledge from biomedical publications in a self-supervised manner alleviates the significant data acquisition and annotation costs across multiple applications in drug development.

\subsection{Instruction Construction}
To make Y-Mol applicable to drug development as a foundational model, we designed multiple types of instructions to finetune it.
\subsubsection{Instructions from Molecule-text Pairs.}
The descriptions from drug databases are easy-to-understand natural language representations of a drug's properties, structure, and function, which contain the rich semantic information that facilitates the alignment between humans and LLMs for the perception of the drug entity. To achieve this, Y-Mol constructs a set of instructions from molecule-text pairs and expert synthetic data to align the human and LLMs. Specifically, we primarily extract molecule-text pairs from DrugBank. The descriptions from DrugBank contain drug-related background knowledge such as symptom-specific therapeutic effects and simple mechanism of action descriptions, which can promote the Y-Mol's understanding of drugs and are beneficial to downstream tasks. Given a pair of molecule-text $(d, s)$, we construct instructions by giving a question ``\textit{Please describe the molecule}: $d_{smiles}$.'' and the corresponding text $s$ is considered the answer, where $d_{smiles}$ denotes the SMILES sequence of $d$. Finally, we capture a set of description-based instructions for finetuning Y-Mol.

\begin{figure}[t]
\centering %
\includegraphics[width=1\columnwidth]{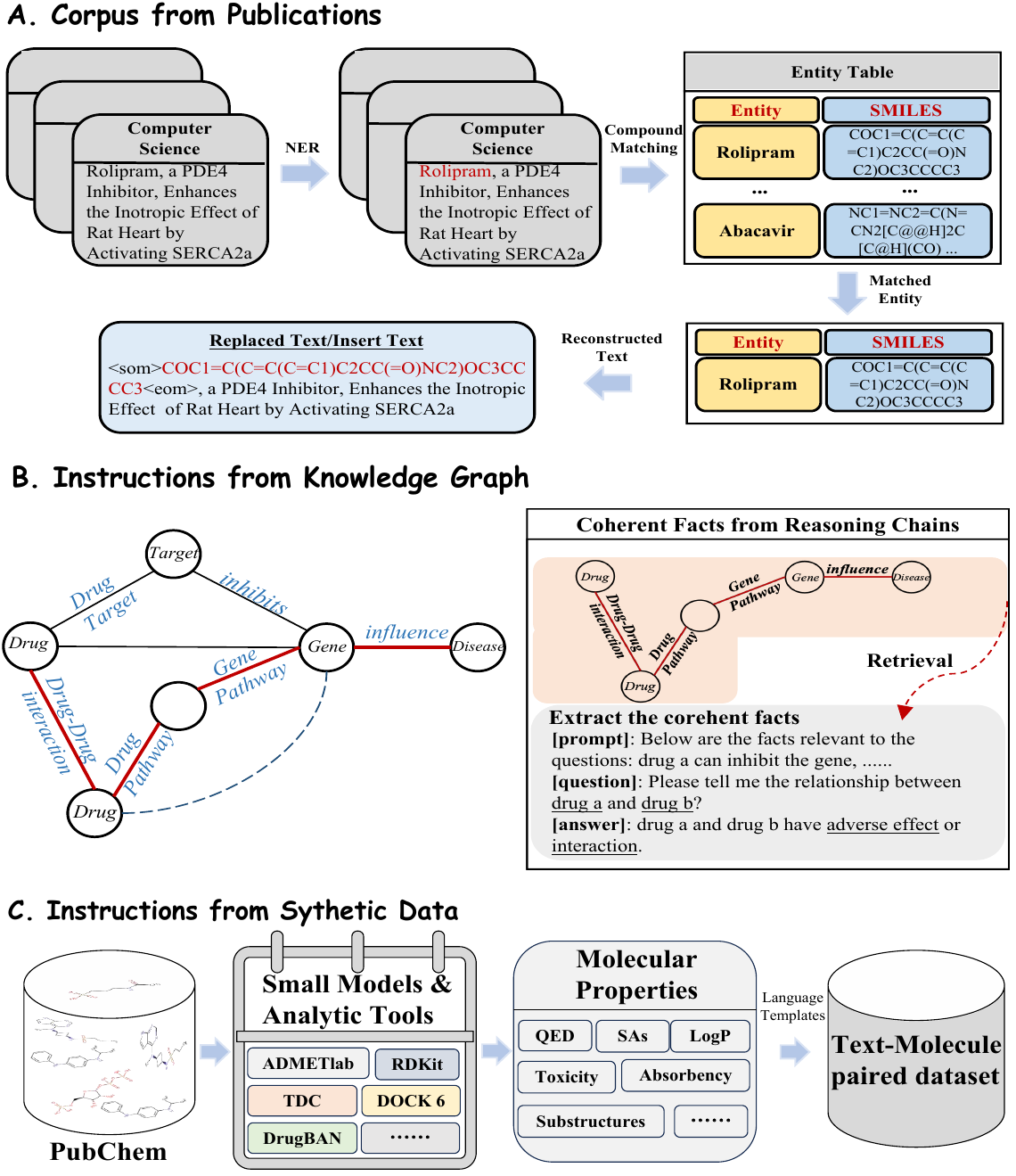} %
\caption{The process of biomedical corpus and instructions: (A) Collecting large-scale biomedical corpus from biomedical publications within the domain of drug discovery. (B) Constructing instructions from coherent facts for enhancing the context of drug-related interactions. (C) Building instructions from expert synthetic data from existing small models to distill knowledge spectrum of drugs into Y-Mol.}
\label{fig:dataset}
\end{figure}
\subsubsection{Instructions from Knowledge Graph.}
Knowledge graphs are structural and encompass extensive semantic descriptions of biological entities, which can effectively improve the performance of LLMs on drug-related tasks (e.g., drug repurposing). To effectively feed the domain knowledge from biomedical KGs, we prompt the facts included in the KGs as natural language. As depicted in Figure~\ref{fig:dataset}B, we iterate over each fact within KGs and extract the enclosing subgraph~\cite{ma2023learning,teru2020inductive} surrounding the target link as the context. After obtaining the enclosing subgraphs (i.e., coherent reasoning chains) of all facts, we prompt them using the pre-designed templates. Specifically, we argue every reasoning chain within the subgraphs has clear relational semantics, thus we extract each coherent path and adopt a well-designed template to convert it into a natural language description as prompt contexts. The constructed contexts combined with the corresponding question are input into Y-Mol and output the supervised answer. Specifically, given a fact $(h,r,t)$ sampled from biomedical KGs, we firstly construct a QA pair as a sample to supervised tuning Y-Mol. For example, we can transfer the link $(h,r,t)$ into the question ``\textit{What is the relationship between $[\mathrm{Entity Type}]$ $h$ and $[\mathrm{Entity Type}]$ $t$}?'' and the answer ``\textit{The $[\mathrm{Entity Type}]$ $h$ plays a role $r$ to the $[\mathrm{Entity Type}]$ $t$.}'', where $[\mathrm{Entity Type}]$ is the generic category attributes (e.g., drug, gene, and disease). To enhance the prediction ability of Y-Mol based on these data, we add the contexts surrounding the target fact as prompts of the QA pairs. Specifically, we extract each path from the enclosing subgraph of the target link and generate a natural language description as follows.
\begin{card}[note] [Example of instructions from KGs]
\textbf{[Prompt]} Below are the facts relevant to the questions:
The drug $a$ can treat disease $b$, and the disease $b$ can be regulated by the gene $c$,...

\textbf{[Question]} What is the relationship between drug $h$ and disease $t$?

\textbf{[Answer]} Drug $h$ plays a role $r$ in the process of $t$.
\end{card}

\subsubsection{Instructions from Expert Synthetic Data.}
In the field of drug development, many computational methods and tools are proposed to solve specific tasks, achieving superior performance. However, there is no standard paradigm to format the overflow of drug development and fully use these existing methods, which limits the performance of computer-assisted drug development. To address this limitation, we introduce Y-Mol to distill existing computational models into LLMs. Specifically, given a drug $d$, to extract more types of molecular properties, we collected a series of molecular tools and advanced models (such as ADMETlab~\cite{fu2024admetlab}, RDKit~\cite{bento2020open}, TDC~\cite{huang2021therapeutics}, DrugBAN~\cite{bai2023interpretable}) to extract molecules with diverse properties (such as QED, SAs, LogP, Toxicity, Absorbency, Substructures) from publicly available datasets\footnote{The adopted properties and datasets are detailed in the Appendix B.1}. As shown in Figure~\ref{fig:dataset}C, Y-Mol is based on the concept of knowledge distillation, where various models and tools are referred to as the ``teacher'', while our model acts as the ``student''. In this way, we can collect the latest models and tools and then use their predicted data to train our model, so our model can evolve in real time. By making use of Y-Mol, we can easily extend our approach to various tasks within drug development. To achieve this, we construct informative instructions based on the inputs and outputs calculated from existing models. An example of a question and answer is shown below.
\begin{card}[note] [Instruction template from Existing Models]
\textbf{[Question]} What is the function of molecule CC(=O)CC?

\textbf{[Answer]} It has a toxicity of 12, an activity of 1.2, a Logp of 3.1, and it is highly volatile.
\end{card}
\noindent In contrast, we can also build reverse instructions by adopting the outputs to construct questions and the inputs as answers to drug design based on specific constraints.

\subsection{Supervised Finetuning}
\begin{figure}[t]
\centering %
\includegraphics[width=0.9\columnwidth]{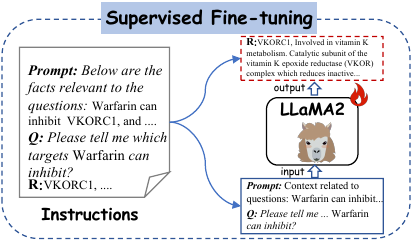} %
\caption{The process of supervised finetuning of Y-Mol based on designed instructions.}
\label{fig:sft}
\end{figure}

Supervised finetuning (SFT) is a technique used to adapt a pretrained LLM to a specific downstream task using labeled data. As shown in Figure~\ref{fig:sft}, we adopt the generated instructions as the supervised inputs and feed them into Y-Mol to finetune a well-performance LLM. Specifically, we input the constructed prompt contexts and questions into Y-Mol and utilize the built answers to supervise the generated outputs from it. After finetuning Y-Mol based on constructed instructions, we employ it in downstream tasks across lead compound discovery, pre-clinic, and clinic predictions.

\subsection{Downstream Tasks}
To validate the effectiveness of our proposed Y-Mol for drug development, we design various tasks across lead compound discovery, pre-clinic, and clinic predictions. Specifically, we introduce four groups of tasks to evaluate it from different stages of drug development: (1) \textbf{Virtual Screening} and \textbf{Drug Design} for lead compound discovery; (2) \textbf{Property Prediction} for identifying physical and chemical properties of discovered lead compounds in the pre-clinic stage; (3) \textbf{Drug Interaction Prediction} for predicting potential adverse drug events in clinic stage. Below we detail these downstream tasks in the context of Y-Mol.

\subsubsection{Virtual Screening.}
Virtual screening~\cite{ahmed2023drug} refers to the search for potential new therapeutic effects from known drugs or targets. This usually involves the prediction of drug-target interaction~\cite{ma2023learning,ma2022kg,pan2022deep}. In this paper, we consider drug-target interaction (DTI) prediction as a downstream task to validate the performance of discovering lead compounds. Specifically, given a drug $d$ and a target (i.e., protein/gene) $p$, we construct a query to predict whether the drug $d$ can interact with the target $p$ based on Y-Mol. For example, given the drug-target pair (\textit{Acetadote}, \textit{Glutathione synthetase}), we create a query as follows: 
\begin{card}[note] [Queray of Virtual Screening] \label{card_vs}
\textbf{[Context]} Below are the facts relevant to the questions: The drug \textit{Acetadote} can..., and the target \textit{Glutathione synthetase} can inhibit...;

\textbf{[Question]} Does the drug \textit{Acetadote} have interaction with the target \textit{Glutathione synthetase}?
\end{card}
\noindent\textbf{Drug Design.}
Drug design aims to generate effective compounds for specific conditions (e.g., the target of disease). In this paper, we view the drug design as a generation task by creating efficient molecular SMILES from Y-Mol. The generated molecules can be adopted as lead compounds in the discovery stage. Specifically, given a target (i.e., the condition) and a paragraph of description as a query, Y-Mol generates a target molecule with SMILES sequence from the context of the query. The generated SMILES can be considered as the designed target molecules. Note that this description can be a constraint on the molecules to be generated in terms of their properties (e.g., LogP and drug-likeness) and functions (e.g., antimicrobial).

\subsubsection{Property Prediction.}
Property prediction of molecules is a fundamental task for drug development in the pre-clinic stage. In this paper, we focus on the physical and chemical properties that are crucial to drug discovery. To fully consider the drug-related expert knowledge, we adopt the available expert small models (e.g., ADMETlab) to infer possible properties and distill this knowledge into Y-Mol. Given a molecule and the property to be predicted, we construct a query as a line of natural description to ask Y-Mol the possible answers.

\subsubsection{Drug Interaction Prediction.}
Identifying potential drug interaction events is a key task for safe drug use for humans in the clinical stage. Similar to the virtual screening, we primarily use the coherent reasoning chains within the biomedical knowledge graph to extract the context of the corresponding drug interaction. The extracted context is adopted to prompt the question ``\textit{What are the side effects for humans taking both drugs $a$ and $b$?}'' as a query. Subsequently, we feed the query into Y-Mol and ask the specific answers.

\section{Experiments}
In this section, we carefully design some key experiments to evaluate the proposed Y-Mol.

\begin{figure}
\centering %
\includegraphics[width=0.8\columnwidth]{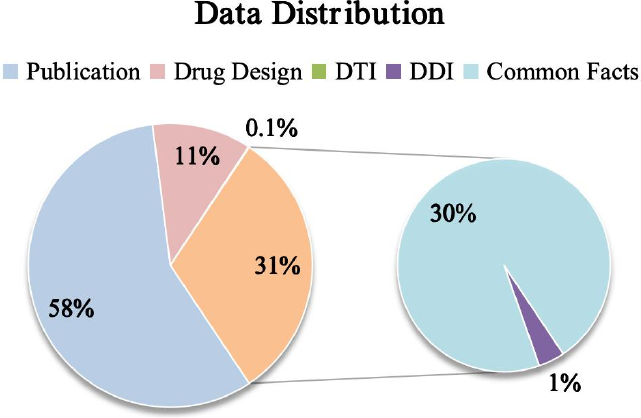} %
\caption{The data distribution of Y-Mol in pretraining and supervised fine-tuning stages across different tasks.}
\label{fig:data_dist}
\end{figure}

\begin{table}[t]
\centering

\resizebox{\columnwidth}{!}{
\begin{tabular}{llrl} \toprule
\textbf{Task}                   & \textbf{Dataset} & \multicolumn{1}{l}{\textbf{Y-Mol}} & \textbf{LLaMA2} \\ \midrule
\multirow{2}{*}{DTI prediction} & DrugBank         & 0.8199                                & 0.7697               \\
                                & Drugcentral      & 0.8331                                & 0.7918               \\
\multirow{2}{*}{DDI prediction} & Ryu's dataset    & 0.6523                                & 0.5031               \\
                                & Deng's dataset   & 0.6219                                & 0.4973       \\ \bottomrule  
                                
\end{tabular}
}
\caption{The performance comparison (ROC-AUC) of DTI and DDI prediction on four datasets.}
\label{tab:virtual}
\end{table}

\subsection{Datasets}
\noindent\textbf{Training.} In this paper, we pre-train Y-Mol using two types of datasets: (1) biomedical text corpus of publications and (2) supervised instructions constructed from biomedical knowledge graphs, and inference data from expert models. For the biomedical corpus, we mine the abstract text and brief introduction from online publications\footnote{\url{https://pubmed.ncbi.nlm.nih.gov/}}. To construct high-quality instructions, we extract the context of specific facts as queries from biomedical knowledge graphs Hetionet~\cite{himmelstein2017systematic} and DRKG~\cite{drkg2020}. In addition, to collect large-scale instructions based on drug-related properties and domain knowledge, we infer the input-output pairs from expert small models to build question-answer pairs. Finally, we collect $\sim11.2$M corpus and $\sim2.3$M instructions. We show the data distribution of Y-Mol across different tasks in Figure~\ref{fig:data_dist}. 

\noindent\textbf{Inference.} To evaluate the performance of Y-Mol on drug-target interaction (DTI) prediction and drug-drug interaction (DDI) prediction, we adopt the widely-used benchmarks \textit{DrugBank}~\cite{knox2024drugbank} and \textit{DrugCentral}~\cite{avram2023drugcentral} for DTI prediction. Meanwhile, we utilize \textit{Ryu's}~\cite{ryu2018deep} and \textit{Deng's}~\cite{deng2020multimodal} datasets to assess DDI prediction.

\subsection{Evaluation}
We design various downstream tasks to evaluate the effectiveness of Y-Mol
in drug development. Specifically, we view drug-target prediction (i.e., virtual screening) as a \textit{Yes or No} question for Y-Mol, which is similar to a binary classification task. Similarly, we consider the drug-drug interaction prediction as a classification task, identifying whether the interaction exists or not. Therefore, we adopt the \textit{receiver operating characteristic curve} (ROC-AUC) to evaluate Y-Mol. In addition, we model the property predictions as regression tasks, which predict the value of specific properties. We utilize \textit{R-square} ($R^2$) to assess the prediction capability of Y-Mol. In contrast, the drug design is considered a generation task and we adopt the standard metrics Valid, Unique, Novelty, and Diversity to evaluate the Y-Mol. To evaluate Y-Mol, we split DTIs and DDIs in each dataset into training and test sets with a ratio of 9:1. For the evaluation of drug design, we randomly sample 200,000 data to assess the generation quality of drugs with single- or multiple-objective. The detailed evaluation settings of drug design can be found in Appendix B.3. Meanwhile, we carefully conduct experiments to measure the predictive performance of 12 chemical and physical properties. For every property, we split the corresponding data into training and test sets with a 9:1 ratio. For a fair comparison, we adopt LLaMA2-7b as the baseline method. More details can be found in Appendix B.2.


\subsection{Main Results}
\noindent\textbf{DTI prediction.}
Identifying unknown drug-target interactions is crucial to virtual screening. As shown in Table~\ref{tab:virtual}, our proposed Y-Mol outperforms LLaMA2 with an improvement of 5.02\% and 4.13\% on the AUC score over DrugBank and DrugCentral datasets, respectively. The results demonstrate that Y-Mol with supervised tuning for biomedical knowledge across multiscale data sources has a positive enhancement on DTI prediction. This shows that Y-Mol has a superior performance for virtual screening.


\begin{figure*}[t]
\centering %
\includegraphics[width=1\textwidth]{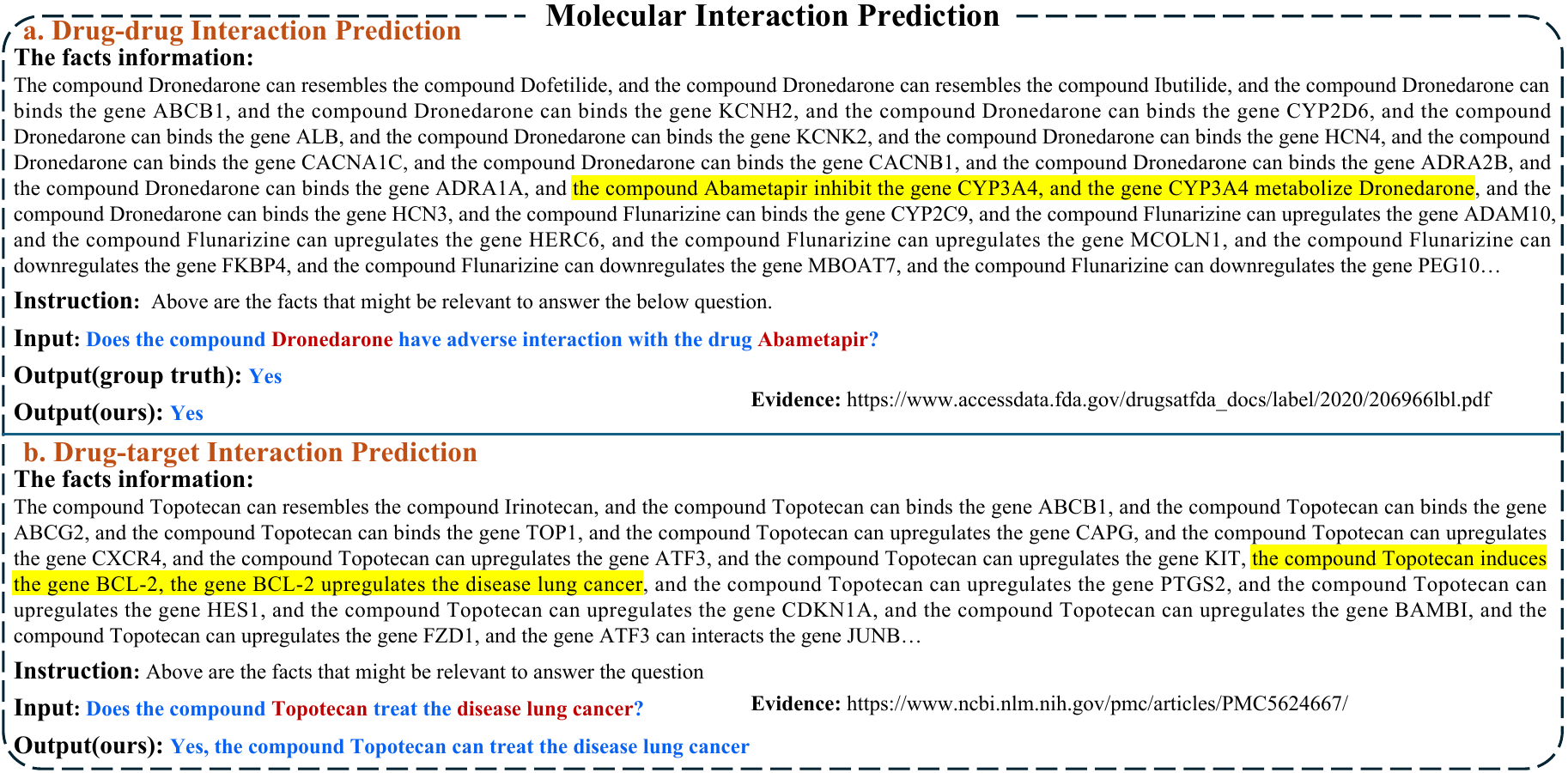} %
\caption{The cases of drug-drug interaction (DDI) prediction and drug-target interaction (DTI) prediction based on Y-Mol.}
\label{fig:case_sf}
\end{figure*}

\begin{table}[t]
\resizebox{\columnwidth}{!}{
\begin{tabular}{lrrrr} \toprule
\multicolumn{1}{c}{\textbf{Single Objective}}   & \multicolumn{1}{l}{Valid} & \multicolumn{1}{l}{Unique} & \multicolumn{1}{l}{Novelty} & \multicolumn{1}{l}{Diversity} \\ \midrule
BBB                                             & 1                         & 0.999                      & 0.751                       & 0.921                         \\
LogP                                            & 0.997                     & 0.999                      & 0.39                        & 0.913                         \\
QED                                             & 1                         & 0.2                        & 0.46                        & 0.879                         \\
SAs                                             & 0.998                     & 0.999                      & 0.502                       & 0.917                         \\
IsValid                                         & 1                         & 0.165                      & 0.68                        & 0.821                         \\ \midrule
\multicolumn{1}{c}{\textbf{Multiple Objective}} & \multicolumn{1}{l}{Valid} & \multicolumn{1}{l}{Unique} & \multicolumn{1}{l}{Novelty} & \multicolumn{1}{l}{Diversity} \\ \midrule
IsValid, BBB, QED                               & 0.997                     & 0.999                      & 0.815                       & 0.913                         \\
IsValid, BBB, QED, SAs                          & 1                         & 0.999                      & 0.821                       & 0.911                         \\
IsValid, LogP                                   & 0.999                     & 0.999                      & 0.522                       & 0.911                         \\
IsValid, BBB                                    & 0.993                     & 0.999                      & 0.872                       & 0.917                         \\
IsValid, QED                                    & 0.998                     & 0.994                      & 0.664                       & 0.913         \\ \bottomrule               
\end{tabular}}
\caption{The performance of Y-Mol for drug design.}
\label{tab:drug_design}
\end{table}

\begin{figure}
\centering %
\includegraphics[width=0.95\columnwidth]{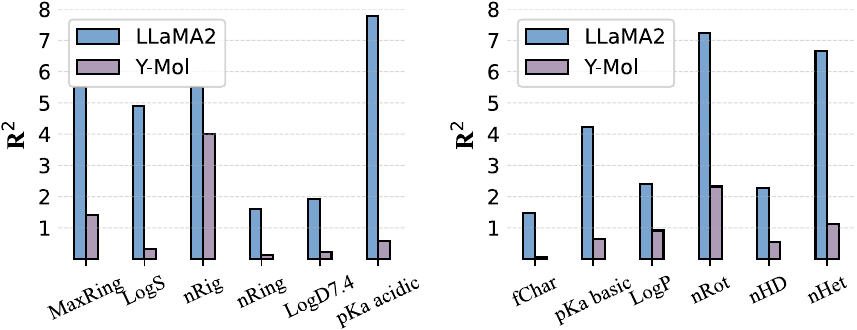} %
\caption{The performance of Y-Mol for physicochemical properties (including 12 types) prediction.}
\label{fig:property}
\end{figure}
\noindent\textbf{Drug Design.}
To validate the performance of Y-Mol in discovering novel lead compounds, we introduce the drug design task. As depicted in Table~\ref{tab:drug_design}, the overall performance of Y-Mol achieves a superior level. In contrast, the LLaMA2-7b can not generate valid molecules with a bad capacity for domain adaptation\footnote{Results generated from LLaMA2-7b in Appendix C.1}. As mentioned in \textbf{Section Method}, the instructions constructed from expert models can introduce multiple constraints for drug design. Therefore, we simultaneously test the drug design performance of Y-Mol across multiple objectives (e.g., LogP and QED). Y-Mol also performs well in this scenario.

\noindent\textbf{Property Prediction.}
To assess the ability of Y-Mol to predict the chemical and physical properties of molecules, we design a molecular property prediction task. The performance is shown in Figure~\ref{fig:property}, we can observe that Y-Mol has a better $R^2$ score (the smaller, the better) than LLaMA2 on all tasks, indicating Y-Mol has a superior generability on predicting chemical and physical properties. This phenomenon proves that Y-Mol is comparable in pre-clinic predictions.

\noindent\textbf{Drug Interaction Prediction.}
In the clinical stage of drug development, predicting potential drug-drug interactions is a key task for ensuring the safe use of drugs. The predictive performance is depicted in Table~\ref{tab:virtual}, we can observe that Y-Mol achieves superior performance on identifying potential drug interaction events. This indicates that Y-Mol learning biomedical knowledge is beneficial to clinic predictions.

\begin{figure}
\centering %
\includegraphics[width=0.9\columnwidth]{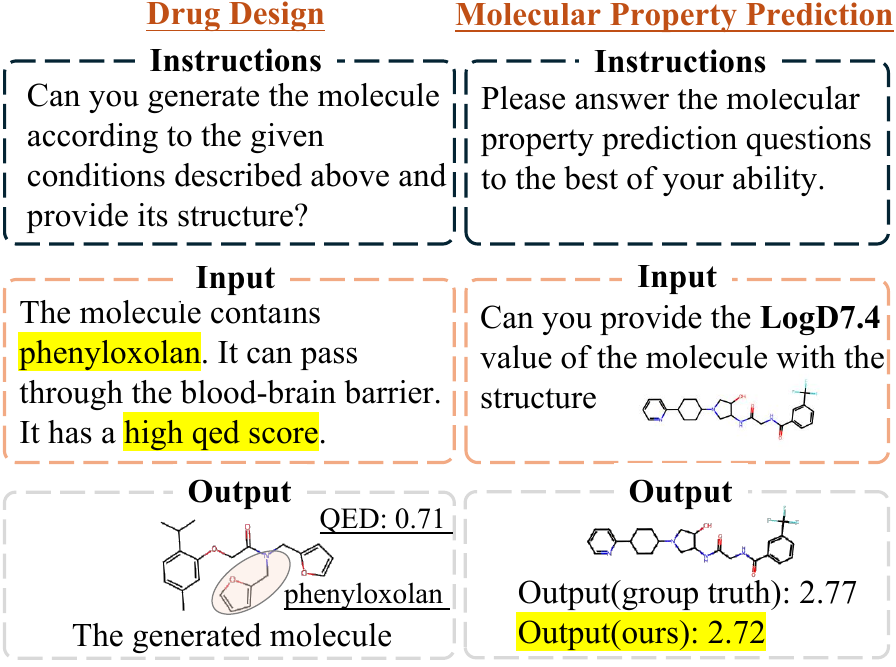} %
\caption{Case study on drug design and molecular physicochemical property prediction.}
\label{fig:case_drug_design}
\end{figure}

\noindent\textbf{Case Study.}
To further evaluate Y-Mol, we show some cases for each task of drug development in Figure~\ref{fig:case_sf} and Figure~\ref{fig:case_drug_design}. For the adverse interaction between \textit{Dronedarone} and \textit{Abametapir}, Y-Mol can find \textit{Abametapir} inhibit the target \textit{CYP2C9} that metabolizes \textit{Dronedarone}, resulting in the serum concentration of \textit{Dronedarone} can be increased when it is combined with \textit{Abametapir}. In addition, the designed drug effectively satisfies the proposed constraints in the query. Similarly, Y-Mol predicts the LogD7.4 of a given molecule accurately with a minor gap. These cases show that Y-Mol is effective in solving tasks of drug development.

\section{Conclusion}
We proposed a multiscale knowledge-guided LLM to build a paradigm for drug development, called Y-Mol. To align the perceptions of drug development between experts and Y-Mol, we build a large-scale biomedical corpus from publications and instructions to distill the drug-related knowledge spectrum into the LLM, alleviating the data acquisition and annotating. 
Experiments demonstrate Y-Mol outperforms LLaMA2-7b in various tasks. In future work, we will promote Y-Mol to the cellular expression level.


\bibliography{main_aaai25}

\begin{thebibliography}{44}
\providecommand{\natexlab}[1]{#1}

\bibitem[{Ahmed et~al.(2023)Ahmed, Kang, Kim, Asif, Rahim, Samantasinghar, Memon, and Choi}]{ahmed2023drug}
Ahmed, F.; Kang, I.~S.; Kim, K.~H.; Asif, A.; Rahim, C. S.~A.; Samantasinghar, A.; Memon, F.~H.; and Choi, K.~H. 2023.
\newblock Drug repurposing for viral cancers: A paradigm of machine learning, deep learning, and virtual screening-based approaches.
\newblock \emph{Journal of Medical Virology}, 95(4): e28693.

\bibitem[{Ajagekar and You(2023)}]{ajagekar2023molecular}
Ajagekar, A.; and You, F. 2023.
\newblock Molecular design with automated quantum computing-based deep learning and optimization.
\newblock \emph{Npj Computational Materials}, 9(1): 143.

\bibitem[{Avram et~al.(2023)Avram, Wilson, Curpan, Halip, Borota, Bora, Bologa, Holmes, Knockel, Yang et~al.}]{avram2023drugcentral}
Avram, S.; Wilson, T.~B.; Curpan, R.; Halip, L.; Borota, A.; Bora, A.; Bologa, C.~G.; Holmes, J.; Knockel, J.; Yang, J.~J.; et~al. 2023.
\newblock DrugCentral 2023 extends human clinical data and integrates veterinary drugs.
\newblock \emph{Nucleic Acids Research}, 51(D1): D1276--D1287.

\bibitem[{Bai et~al.(2023)Bai, Miljkovi{\'c}, John, and Lu}]{bai2023interpretable}
Bai, P.; Miljkovi{\'c}, F.; John, B.; and Lu, H. 2023.
\newblock Interpretable bilinear attention network with domain adaptation improves drug--target prediction.
\newblock \emph{Nature Machine Intelligence}, 5(2): 126--136.

\bibitem[{Bento et~al.(2020)Bento, Hersey, F{\'e}lix, Landrum, Gaulton, Atkinson, Bellis, De~Veij, and Leach}]{bento2020open}
Bento, A.~P.; Hersey, A.; F{\'e}lix, E.; Landrum, G.; Gaulton, A.; Atkinson, F.; Bellis, L.~J.; De~Veij, M.; and Leach, A.~R. 2020.
\newblock An open source chemical structure curation pipeline using RDKit.
\newblock \emph{Journal of Cheminformatics}, 12: 1--16.

\bibitem[{Chaves et~al.(2024)Chaves, Wang, Tu, Vaishnav, Lee, Mahdavi, Semturs, Fleet, Natarajan, and Azizi}]{chaves2024tx}
Chaves, J. M.~Z.; Wang, E.; Tu, T.; Vaishnav, E.~D.; Lee, B.; Mahdavi, S.~S.; Semturs, C.; Fleet, D.; Natarajan, V.; and Azizi, S. 2024.
\newblock Tx-LLM: A Large Language Model for Therapeutics.
\newblock \emph{arXiv preprint arXiv:2406.06316}.

\bibitem[{Deng et~al.(2020)Deng, Xu, Qiu, Xia, Zhang, and Liu}]{deng2020multimodal}
Deng, Y.; Xu, X.; Qiu, Y.; Xia, J.; Zhang, W.; and Liu, S. 2020.
\newblock A multimodal deep learning framework for predicting drug--drug interaction events.
\newblock \emph{Bioinformatics}, 36(15): 4316--4322.

\bibitem[{Edfeldt et~al.(2024)Edfeldt, Edwards, Engkvist, G{\"u}nther, Hartley, Hulcoop, Leach, Marsden, Menge, Misquitta et~al.}]{edfeldt2024data}
Edfeldt, K.; Edwards, A.~M.; Engkvist, O.; G{\"u}nther, J.; Hartley, M.; Hulcoop, D.~G.; Leach, A.~R.; Marsden, B.~D.; Menge, A.; Misquitta, L.; et~al. 2024.
\newblock A data science roadmap for open science organizations engaged in early-stage drug discovery.
\newblock \emph{Nature Communications}, 15(1): 5640.

\bibitem[{Edwards et~al.(2022)Edwards, Lai, Ros, Honke, Cho, and Ji}]{edwards-etal-2022-translation}
Edwards, C.; Lai, T.; Ros, K.; Honke, G.; Cho, K.; and Ji, H. 2022.
\newblock Translation between Molecules and Natural Language.
\newblock In \emph{Proceedings of the 2022 Conference on Empirical Methods in Natural Language Processing}, 375--413. Abu Dhabi, United Arab Emirates: Association for Computational Linguistics.

\bibitem[{Fang et~al.(2023)Fang, Liang, Zhang, Liu, Huang, Chen, Fan, and Chen}]{fang2023mol}
Fang, Y.; Liang, X.; Zhang, N.; Liu, K.; Huang, R.; Chen, Z.; Fan, X.; and Chen, H. 2023.
\newblock Mol-instructions: A large-scale biomolecular instruction dataset for large language models.
\newblock \emph{arXiv preprint arXiv:2306.08018}.

\bibitem[{Fu et~al.(2024)Fu, Shi, Yi, Wang, He, Wu, Peng, Deng, Wang, Wu et~al.}]{fu2024admetlab}
Fu, L.; Shi, S.; Yi, J.; Wang, N.; He, Y.; Wu, Z.; Peng, J.; Deng, Y.; Wang, W.; Wu, C.; et~al. 2024.
\newblock ADMETlab 3.0: an updated comprehensive online ADMET prediction platform enhanced with broader coverage, improved performance, API functionality and decision support.
\newblock \emph{Nucleic Acids Research}, gkae236.

\bibitem[{Himmelstein et~al.(2017)Himmelstein, Lizee, Hessler, Brueggeman, Chen, Hadley, Green, Khankhanian, and Baranzini}]{himmelstein2017systematic}
Himmelstein, D.~S.; Lizee, A.; Hessler, C.; Brueggeman, L.; Chen, S.~L.; Hadley, D.; Green, A.; Khankhanian, P.; and Baranzini, S.~E. 2017.
\newblock Systematic integration of biomedical knowledge prioritizes drugs for repurposing.
\newblock \emph{Elife}, 6: e26726.

\bibitem[{Huang et~al.(2021)Huang, Fu, Gao, Zhao, Roohani, Leskovec, Coley, Xiao, Sun, and Zitnik}]{huang2021therapeutics}
Huang, K.; Fu, T.; Gao, W.; Zhao, Y.; Roohani, Y.; Leskovec, J.; Coley, C.~W.; Xiao, C.; Sun, J.; and Zitnik, M. 2021.
\newblock Therapeutics data commons: Machine learning datasets and tasks for drug discovery and development.
\newblock \emph{arXiv preprint arXiv:2102.09548}.

\bibitem[{Ioannidis et~al.(2020)Ioannidis, Song, Manchanda, Li, Pan, Zheng, Ning, Zeng, and Karypis}]{drkg2020}
Ioannidis, V.~N.; Song, X.; Manchanda, S.; Li, M.; Pan, X.; Zheng, D.; Ning, X.; Zeng, X.; and Karypis, G. 2020.
\newblock DRKG - Drug Repurposing Knowledge Graph for Covid-19.
\newblock \url{https://github.com/gnn4dr/DRKG/}.

\bibitem[{Knox et~al.(2024)Knox, Wilson, Klinger, Franklin, Oler, Wilson, Pon, Cox, Chin, Strawbridge et~al.}]{knox2024drugbank}
Knox, C.; Wilson, M.; Klinger, C.~M.; Franklin, M.; Oler, E.; Wilson, A.; Pon, A.; Cox, J.; Chin, N.~E.; Strawbridge, S.~A.; et~al. 2024.
\newblock DrugBank 6.0: the DrugBank knowledgebase for 2024.
\newblock \emph{Nucleic Acids Research}, 52(D1): D1265--D1275.

\bibitem[{Li et~al.(2024)Li, Liu, Fan, Wei, Liu, Tang, and Li}]{li2024empowering}
Li, J.; Liu, Y.; Fan, W.; Wei, X.-Y.; Liu, H.; Tang, J.; and Li, Q. 2024.
\newblock Empowering molecule discovery for molecule-caption translation with large language models: A chatgpt perspective.
\newblock \emph{IEEE Transactions on Knowledge and Data Engineering}.

\bibitem[{Liu et~al.(2023{\natexlab{a}})Liu, Nie, Wang, Lu, Qiao, Liu, Tang, Xiao, and Anandkumar}]{liu2023moleculestm}
Liu, S.; Nie, W.; Wang, C.; Lu, J.; Qiao, Z.; Liu, L.; Tang, J.; Xiao, C.; and Anandkumar, A. 2023{\natexlab{a}}.
\newblock Multi-modal molecule structure-text model for text-based retrieval and editing.
\newblock \emph{Nature Machine Intelligence}, 5(12): 1447--1457.

\bibitem[{Liu et~al.(2024)Liu, Ding, Zhou, Fan, and Tan}]{liu2024moleculargpt}
Liu, Y.; Ding, S.; Zhou, S.; Fan, W.; and Tan, Q. 2024.
\newblock MolecularGPT: Open Large Language Model (LLM) for Few-Shot Molecular Property Prediction.
\newblock \emph{arXiv preprint arXiv:2406.12950}.

\bibitem[{Liu et~al.(2023{\natexlab{b}})Liu, Zhang, Xia, Wu, Xie, Qin, Zhang, and Liu}]{liu-etal-2023-molxpt}
Liu, Z.; Zhang, W.; Xia, Y.; Wu, L.; Xie, S.; Qin, T.; Zhang, M.; and Liu, T.-Y. 2023{\natexlab{b}}.
\newblock {M}ol{XPT}: Wrapping Molecules with Text for Generative Pre-training.
\newblock In Rogers, A.; Boyd-Graber, J.; and Okazaki, N., eds., \emph{Proceedings of the 61st Annual Meeting of the Association for Computational Linguistics (Volume 2: Short Papers)}, 1606--1616. Toronto, Canada: Association for Computational Linguistics.

\bibitem[{Luo et~al.()Luo, Ning, Zhao, Wang, Ding, Chen, Fu, Han, Xu, Qiu, Pan, Li, Li, Feng, Tu, Liu, Yang, Wang, Sun, and Lin}]{10.1093/jamia/ocae037}
Luo, L.; Ning, J.; Zhao, Y.; Wang, Z.; Ding, Z.; Chen, P.; Fu, W.; Han, Q.; Xu, G.; Qiu, Y.; Pan, D.; Li, J.; Li, H.; Feng, W.; Tu, S.; Liu, Y.; Yang, Z.; Wang, J.; Sun, Y.; and Lin, H. ????
\newblock {Taiyi: a bilingual fine-tuned large language model for diverse biomedical tasks}.
\newblock \emph{Journal of the American Medical Informatics Association}, ocae037.

\bibitem[{Lyu et~al.(2023)Lyu, Jiang, Zeng, Xia, and Luo}]{lyu2023llm}
Lyu, H.; Jiang, S.; Zeng, H.; Xia, Y.; and Luo, J. 2023.
\newblock Llm-rec: Personalized recommendation via prompting large language models.
\newblock \emph{arXiv preprint arXiv:2307.15780}.

\bibitem[{Ma et~al.(2023)Ma, Chen, Tao, Zheng, Lin, Pang, Liu, Wang, Song, and Zeng}]{ma2023learning}
Ma, T.; Chen, Y.; Tao, W.; Zheng, D.; Lin, X.; Pang, P. C.-l.; Liu, Y.; Wang, Y.; Song, B.; and Zeng, X. 2023.
\newblock Learning to Denoise Unreliable Interactions for Link Prediction on Biomedical Knowledge Graph.
\newblock \emph{arXiv preprint arXiv:2312.06682}.

\bibitem[{Ma et~al.(2022)Ma, Lin, Song, Philip, and Zeng}]{ma2022kg}
Ma, T.; Lin, X.; Song, B.; Philip, S.~Y.; and Zeng, X. 2022.
\newblock Kg-mtl: knowledge graph enhanced multi-task learning for molecular interaction.
\newblock \emph{IEEE Transactions on Knowledge and Data Engineering}, 35(7): 7068--7081.

\bibitem[{OpenAI(2023)}]{openai2023gpt}
OpenAI, R. 2023.
\newblock Gpt-4 technical report. arxiv 2303.08774.
\newblock \emph{View in Article}, 2(5).

\bibitem[{Ouyang et~al.(2022)Ouyang, Wu, Jiang, Almeida, Wainwright, Mishkin, Zhang, Agarwal, Slama, Ray et~al.}]{ouyang2022training}
Ouyang, L.; Wu, J.; Jiang, X.; Almeida, D.; Wainwright, C.; Mishkin, P.; Zhang, C.; Agarwal, S.; Slama, K.; Ray, A.; et~al. 2022.
\newblock Training language models to follow instructions with human feedback.
\newblock \emph{Advances in neural information processing systems}, 35: 27730--27744.

\bibitem[{Pan et~al.(2022)Pan, Lin, Cao, Zeng, Yu, He, Nussinov, and Cheng}]{pan2022deep}
Pan, X.; Lin, X.; Cao, D.; Zeng, X.; Yu, P.~S.; He, L.; Nussinov, R.; and Cheng, F. 2022.
\newblock Deep learning for drug repurposing: Methods, databases, and applications.
\newblock \emph{Wiley interdisciplinary reviews: Computational molecular science}, 12(4): e1597.

\bibitem[{Pereira et~al.(2021)Pereira, Abbasi, Oliveira, Ribeiro, and Arrais}]{pereira2021optimizing}
Pereira, T.; Abbasi, M.; Oliveira, J.~L.; Ribeiro, B.; and Arrais, J. 2021.
\newblock Optimizing blood--brain barrier permeation through deep reinforcement learning for de novo drug design.
\newblock \emph{Bioinformatics}, 37(Supplement\_1): i84--i92.

\bibitem[{Ryu, Kim, and Lee(2018)}]{ryu2018deep}
Ryu, J.~Y.; Kim, H.~U.; and Lee, S.~Y. 2018.
\newblock Deep learning improves prediction of drug--drug and drug--food interactions.
\newblock \emph{Proceedings of the National Academy of Sciences}, 115(18): E4304--E4311.

\bibitem[{Sanh et~al.(2021)Sanh, Webson, Raffel, Bach, Sutawika, Alyafeai, Chaffin, Stiegler, Scao, Raja et~al.}]{sanh2021multitask}
Sanh, V.; Webson, A.; Raffel, C.; Bach, S.~H.; Sutawika, L.; Alyafeai, Z.; Chaffin, A.; Stiegler, A.; Scao, T.~L.; Raja, A.; et~al. 2021.
\newblock Multitask prompted training enables zero-shot task generalization.
\newblock \emph{arXiv preprint arXiv:2110.08207}.

\bibitem[{Sung et~al.(2022)Sung, Jeong, Choi, Kim, Lee, and Kang}]{sung2022bern2}
Sung, M.; Jeong, M.; Choi, Y.; Kim, D.; Lee, J.; and Kang, J. 2022.
\newblock BERN2: an advanced neural biomedical named entity recognition and normalization tool.
\newblock \emph{Bioinformatics}, 38(20): 4837--4839.

\bibitem[{Taori et~al.(2023)Taori, Gulrajani, Zhang, Dubois, Li, Guestrin, Liang, and Hashimoto}]{taori2023stanford}
Taori, R.; Gulrajani, I.; Zhang, T.; Dubois, Y.; Li, X.; Guestrin, C.; Liang, P.; and Hashimoto, T.~B. 2023.
\newblock Stanford alpaca: An instruction-following llama model.

\bibitem[{Teru, Denis, and Hamilton(2020)}]{teru2020inductive}
Teru, K.; Denis, E.; and Hamilton, W. 2020.
\newblock Inductive relation prediction by subgraph reasoning.
\newblock In \emph{International Conference on Machine Learning}, 9448--9457. PMLR.

\bibitem[{Tinn et~al.(2023)Tinn, Cheng, Gu, Usuyama, Liu, Naumann, Gao, and Poon}]{tinn2023fine}
Tinn, R.; Cheng, H.; Gu, Y.; Usuyama, N.; Liu, X.; Naumann, T.; Gao, J.; and Poon, H. 2023.
\newblock Fine-tuning large neural language models for biomedical natural language processing.
\newblock \emph{Patterns}, 4(4).

\bibitem[{Touvron et~al.(2023)Touvron, Lavril, Izacard, Martinet, Lachaux, Lacroix, Rozi{\`e}re, Goyal, Hambro, Azhar et~al.}]{touvron2023llama}
Touvron, H.; Lavril, T.; Izacard, G.; Martinet, X.; Lachaux, M.-A.; Lacroix, T.; Rozi{\`e}re, B.; Goyal, N.; Hambro, E.; Azhar, F.; et~al. 2023.
\newblock Llama: Open and efficient foundation language models.
\newblock \emph{arXiv preprint arXiv:2302.13971}.

\bibitem[{Wang et~al.(2022)Wang, Kordi, Mishra, Liu, Smith, Khashabi, and Hajishirzi}]{wang2022self}
Wang, Y.; Kordi, Y.; Mishra, S.; Liu, A.; Smith, N.~A.; Khashabi, D.; and Hajishirzi, H. 2022.
\newblock Self-instruct: Aligning language models with self-generated instructions.
\newblock \emph{arXiv preprint arXiv:2212.10560}.

\bibitem[{Wei et~al.(2024)Wei, Allot, Lai, Leaman, Tian, Luo, Jin, Wang, Chen, and Lu}]{wei2024pubtator}
Wei, C.-H.; Allot, A.; Lai, P.-T.; Leaman, R.; Tian, S.; Luo, L.; Jin, Q.; Wang, Z.; Chen, Q.; and Lu, Z. 2024.
\newblock PubTator 3.0: an AI-powered literature resource for unlocking biomedical knowledge.
\newblock \emph{Nucleic Acids Research}, gkae235.

\bibitem[{Xiong et~al.(2021)Xiong, Wu, Yi, Fu, Yang, Hsieh, Yin, Zeng, Wu, Lu et~al.}]{xiong2021admetlab}
Xiong, G.; Wu, Z.; Yi, J.; Fu, L.; Yang, Z.; Hsieh, C.; Yin, M.; Zeng, X.; Wu, C.; Lu, A.; et~al. 2021.
\newblock ADMETlab 2.0: an integrated online platform for accurate and comprehensive predictions of ADMET properties.
\newblock \emph{Nucleic acids research}, 49(W1): W5--W14.

\bibitem[{Zeng et~al.(2022{\natexlab{a}})Zeng, Liu, Du, Wang, Lai, Ding, Yang, Xu, Zheng, Xia et~al.}]{zeng2022glm}
Zeng, A.; Liu, X.; Du, Z.; Wang, Z.; Lai, H.; Ding, M.; Yang, Z.; Xu, Y.; Zheng, W.; Xia, X.; et~al. 2022{\natexlab{a}}.
\newblock Glm-130b: An open bilingual pre-trained model.
\newblock \emph{arXiv preprint arXiv:2210.02414}.

\bibitem[{Zeng et~al.(2020)Zeng, Song, Ma, Pan, Zhou, Hou, Zhang, Li, Karypis, and Cheng}]{zeng2020repurpose}
Zeng, X.; Song, X.; Ma, T.; Pan, X.; Zhou, Y.; Hou, Y.; Zhang, Z.; Li, K.; Karypis, G.; and Cheng, F. 2020.
\newblock Repurpose open data to discover therapeutics for COVID-19 using deep learning.
\newblock \emph{Journal of proteome research}, 19(11): 4624--4636.

\bibitem[{Zeng et~al.(2022{\natexlab{b}})Zeng, Xiang, Yu, Wang, Li, Nussinov, and Cheng}]{zeng2022accurate}
Zeng, X.; Xiang, H.; Yu, L.; Wang, J.; Li, K.; Nussinov, R.; and Cheng, F. 2022{\natexlab{b}}.
\newblock Accurate prediction of molecular properties and drug targets using a self-supervised image representation learning framework.
\newblock \emph{Nature Machine Intelligence}, 4(11): 1004--1016.

\bibitem[{Zeng et~al.(2019)Zeng, Zhu, Liu, Zhou, Nussinov, and Cheng}]{zeng2019deepdr}
Zeng, X.; Zhu, S.; Liu, X.; Zhou, Y.; Nussinov, R.; and Cheng, F. 2019.
\newblock deepDR: a network-based deep learning approach to in silico drug repositioning.
\newblock \emph{Bioinformatics}, 35(24): 5191--5198.

\bibitem[{Zeng et~al.(2022{\natexlab{c}})Zeng, Yao, Liu, and Sun}]{zeng2022deep}
Zeng, Z.; Yao, Y.; Liu, Z.; and Sun, M. 2022{\natexlab{c}}.
\newblock A deep-learning system bridging molecule structure and biomedical text with comprehension comparable to human professionals.
\newblock \emph{Nature communications}, 13(1): 862.

\bibitem[{Zhang et~al.(2023)Zhang, Yu, Yan, Liu, Adhikarla, Fu, Chen, Chen, Zhou, Li et~al.}]{zhang2023biomedgpt}
Zhang, K.; Yu, J.; Yan, Z.; Liu, Y.; Adhikarla, E.; Fu, S.; Chen, X.; Chen, C.; Zhou, Y.; Li, X.; et~al. 2023.
\newblock Biomedgpt: A unified and generalist biomedical generative pre-trained transformer for vision, language, and multimodal tasks.
\newblock \emph{arXiv preprint arXiv:2305.17100}.

\bibitem[{Zhou et~al.(2024)Zhou, Wang, Li, Wang, Liu, Sun, Lin, Wang, and Zeng}]{zhou2024instruction}
Zhou, P.; Wang, J.; Li, C.; Wang, Z.; Liu, Y.; Sun, S.; Lin, J.; Wang, L.; and Zeng, X. 2024.
\newblock Instruction Multi-Constraint Molecular Generation Using a Teacher-Student Large Language Model.
\newblock \emph{arXiv preprint arXiv:2403.13244}.

\end{thebibliography}
\section{Reproducibility Checklist}
This paper
\begin{itemize}
    \item Includes a conceptual outline and/or pseudocode description of AI methods introduced [\textbf{yes}]
    \item Clearly delineates statements that are opinions, hypothesis, and speculation from objective facts and results [\textbf{yes}]
    \item Provides well marked pedagogical references for less-familiare readers to gain background necessary to replicate the paper [\textbf{yes}]
\end{itemize}
Does this paper make theoretical contributions? [\textbf{yes}]

\noindent If yes, please complete the list below.
\begin{itemize}
    \item All assumptions and restrictions are stated clearly and formally. [\textbf{no}] We experimentally prove novel claims rather than theorem statements.
    \item All novel claims are stated formally (e.g., in theorem statements). [\textbf{partial}]
    \item Proof sketches or intuitions are given for complex and/or novel results. [\textbf{yes}]
    \item Appropriate citations to theoretical tools used are given. [\textbf{yes}]
    \item All theoretical claims are demonstrated empirically to hold. [\textbf{NA}]
    \item All experimental code used to eliminate or disprove claims is included. [\textbf{NA}]
\end{itemize}
Does this paper rely on one or more datasets? [\textbf{yes}]

\noindent If yes, please complete the list below.
\begin{itemize}
    \item A motivation is given for why the experiments are conducted on the selected datasets [\textbf{yes}]
    \item All novel datasets introduced in this paper are included in a data appendix. [\textbf{yes}] The details are shown in the Appendix and an anonymous repository.
    \item All novel datasets introduced in this paper will be made publicly available upon publication of the paper with a license that allows free usage for research purposes. [\textbf{yes}]
    \item All datasets drawn from the existing literature (potentially including authors’ own previously published work) are accompanied by appropriate citations. [\textbf{yes}]
    \item All datasets drawn from the existing literature (potentially including authors’ own previously published work) are publicly available. [\textbf{yes}]
    \item All datasets that are not publicly available are described in detail, with explanation why publicly available alternatives are not scientifically satisficing. [\textbf{NA}]
\end{itemize}
Does this paper include computational experiments? [\textbf{yes}]

\noindent If yes, please complete the list below.

\begin{itemize}
    \item Any code required for pre-processing data is included in the appendix. [\textbf{yes}].
    \item All source code required for conducting and analyzing the experiments is included in a code appendix. [\textbf{yes}]
    \item All source code required for conducting and analyzing the experiments will be made publicly available upon publication of the paper with a license that allows free usage for research purposes. [\textbf{yes}]
    \item All source code implementing new methods have comments detailing the implementation, with references to the paper where each step comes from [\textbf{yes}]
    \item If an algorithm depends on randomness, then the method used for setting seeds is described in a way sufficient to allow replication of results. [\textbf{yes}]
    \item This paper specifies the computing infrastructure used for running experiments (hardware and software), including GPU/CPU models; amount of memory; operating system; names and versions of relevant software libraries and frameworks. [\textbf{yes}]
    \item This paper formally describes evaluation metrics used and explains the motivation for choosing these metrics. [\textbf{yes}]
    \item This paper states the number of algorithm runs used to compute each reported result. [\textbf{yes}]
    \item Analysis of experiments goes beyond single-dimensional summaries of performance (e.g., average; median) to include measures of variation, confidence, or other distributional information. [\textbf{yes}]
    \item The significance of any improvement or decrease in performance is judged using appropriate statistical tests (e.g., Wilcoxon signed-rank). [\textbf{Partial}]
    \item This paper lists all final (hyper-)parameters used for each model/algorithm in the paper’s experiments. [\textbf{yes}]
    \item This paper states the number and range of values tried per (hyper-) parameter during development of the paper, along with the criterion used for selecting the final parameter setting. [\textbf{no}]
\end{itemize}

\clearpage

\section{A Details of Processed Data}
\subsection{A.1 Related topics for mining texts from online publications}
To emphasize the key information about drugs, we select the
most related topics to mine meaningful texts. Specifically, we select publications within the areas of \textit{Chemistry}, \textit{Computer Science}, and \textit{Bioinformatics}. The data distributions of them are shown in Table~\ref{tab:appdx_ds}. All datasets are available at \url{https://anonymous.4open.science/r/Y-MOL}

\subsection{A.2 The overall data distribution for training}

In this paper, we design an LLM paradigm for drug development by learning large-scale biomedical knowledge from publications, knowledge graphs (KGs), and expert models or tools. The data distribution is shown in Table~\ref{tab:data_stat}. Y-Mol learning from the large-scale text corpus within publications contains enough background knowledge to facilitate finer-grained finetuning.

\begin{figure*}
\centering %
\includegraphics[width=1\textwidth]{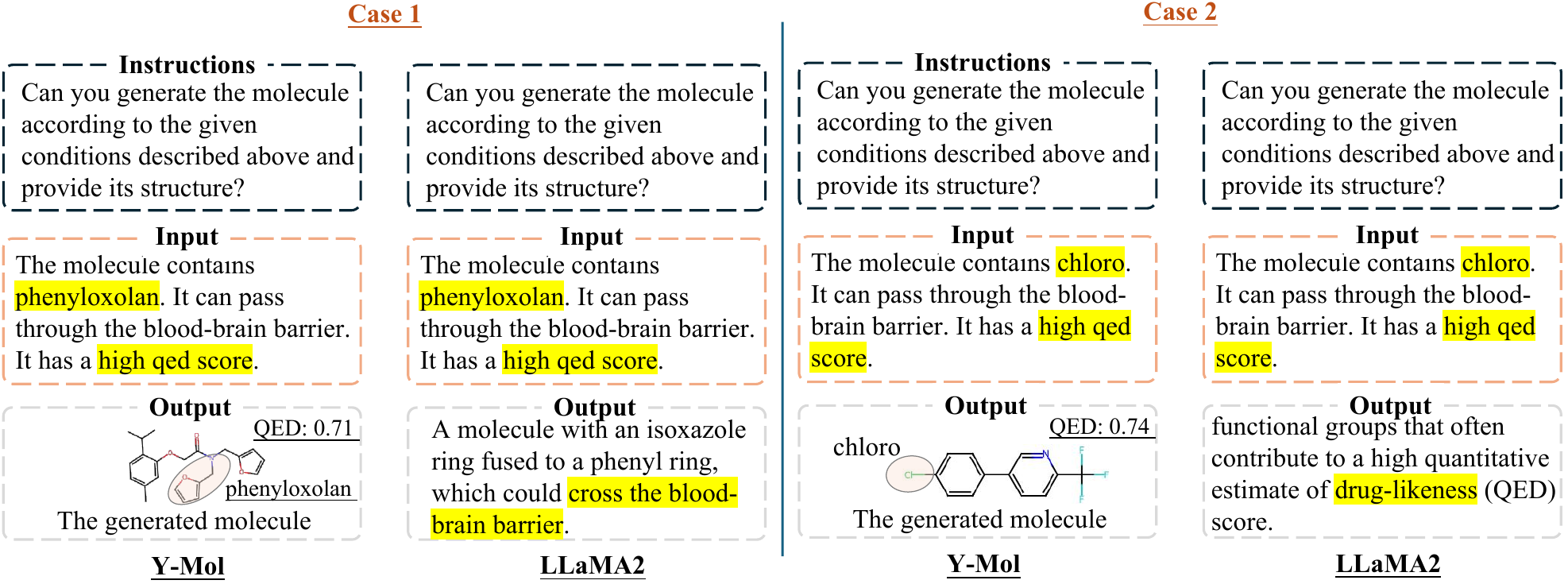} %
\caption{The cases of Y-Mol and LLaMA2-7b for generating drugs. Y-Mol can effectively design drugs, while LLaMA2-7b can not fully understand the question (only knowing the blood-brain barrier and the drug-likeness) and is invalid in generating molecules.}
\label{fig:case_appendx}
\end{figure*}
\section{B Details of Experimental Settings}
\subsection{B.1 Meanings of predicted properties}
\begin{table}[t]
\centering
\begin{tabular}{lrr} \toprule
\textbf{Topic}   & \multicolumn{1}{l}{\textbf{\# Publications}} & \multicolumn{1}{l}{\textbf{Proportion}} \\ \midrule
Chemistry        & 4,237,738                                             & 37.71\%                                 \\
Bioinformatics   & 3,200,657                                             & 28.48\%                                 \\
Computer Science & 3,798,927                                             & 33.81\%              \\ \bottomrule  
\end{tabular}
\caption{The data statistic of publications in various topics.}
\label{tab:appdx_ds}
\end{table}
\begin{table}[t]

\resizebox{\columnwidth}{!}{
\begin{tabular}{llrl} \toprule
\textbf{Data Source}              & \textbf{Task}           & \multicolumn{1}{l}{\textbf{Scale}} & \textbf{Type} \\ \midrule
Publication                            & Text Generation         & 11,237,322                           & Pretrain  \\
                                  & Drug-Target Interaction & 18,462                              & SFT                    \\
                                  & Drug-Drug Interaction   & 245,569                             & SFT                    \\
\multirow{-3}{*}{KG} & Common Facts            & {\color[HTML]{1F2328} 5,874,261}   & SFT                    \\
Expert Models                     & Drug Design             & 2,211,531                            & SFT                \\ \bottomrule   
\end{tabular}
}
\caption{The statistics of used datasets and corresponding tasks.}
\label{tab:data_stat}
\end{table}

\begin{table}[t]
\resizebox{\columnwidth}{!}{
\begin{tabular}{ll} \toprule
\textbf{Properties} & \textbf{Meanings}                                      \\ \midrule
MaxRing             & Refers to the largest ring structure in a molecule     \\
LogS                & Logarithm of Solubility                                \\
nRig                & Number of rigid bonds                                  \\
nRing               & Number of rings in a molecule                          \\
LogD7.4             & Logarithmic of the distribution factor under PH=7.4    \\
pKa acidic          & Negative logarithmic of the acid dissociation constant \\
fChar               & {\color[HTML]{060607} Fingerprint characteristic}      \\
pKa basic           & Negative logarithmic of the base dissociation constant \\
LogP                & Logarithmic of the octanol-water partition coefficient \\
nRot                & Number of rotatable keys                               \\
nHD                 & {\color[HTML]{060607} Hydrogen Bond Donor}             \\
nHet                & Number of heteroatoms   \\
\bottomrule             
\end{tabular}}
\caption{The meanings of predicted properties.}
\label{tab:appdx_mean}
\end{table}
We show the corresponding meanings of properties in the property prediction task in Table~\ref{tab:appdx_mean}. These adopted properties are crucial to drug metabolism in the flow of drug development.

\subsection{B.2 Implementation Details}
In the pretraining stage, we set the $lr=1.2\times10^{-5}$, the number of layers is 32, and the hidden size is 4096. The corresponding config of hyperparameters is available at \url{https://anonymous.4open.science/r/Y-MOL/config.txt}.

During the supervised finetuning process, we set the learning rate to $lr=4\times10^{-5}$ and the number of iterations to 3. The batch size was set to 8. Considering the memory limitations and the size of the model, we employed the gradient accumulation technique by setting the gradient accumulation steps parameter to 4, which accumulates the gradients equivalent to a batch size of 32 per iteration, simulating the effect of training with a larger batch size. We utilized the DeepSpeed library, combined with the ZeRO-2 optimization technique and Bfloat16 (BF16) precision, to achieve efficient memory usage and accelerate the training process while maintaining model performance. The experiments were conducted on an Ubuntu 20.04.3 LTS operating system with CUDA version 11.7. The server was equipped with 4 A800 GPUs, each with 80GB of VRAM, providing ample computational resources for the full finetuning of large models. These configurations ensured optimized memory usage and training speed while preserving model performance.

\subsection{B.3 Details of Drug Design}
\subsubsection{B.3.1 Details of adopted objectives.}
Following~\cite{zhou2024instruction,ajagekar2023molecular,pereira2021optimizing}, we design a multi-constraint drug design under different molecular properties. To evaluate Y-Mol, we introduce two strategies: (1) drug design for a single objective and (2) drug design for multiple objectives. Specifically, we utilize BBB\footnote{BBB is the degree of crossing the blood-brain barrier.}, LogP\footnote{LogP assess the lipophilicity of compounds}, QED\footnote{QED is the drug-likeness, which is key to designing new drugs.}, SAs\footnote{SAs indicates the synthesizability of compounds}, and IsValid\footnote{IsValid indicate whether the SMILES is valid or not.} as our objectives. These objectives are crucial to designing new drugs. As we know, an effective drug usually needs to fulfill all the above objectives. To evaluate Y-Mol deeply, we conduct experiments to design drugs under multiple objectives. The experimental results in the main paper show that Y-Mol is effective enough in drug design.

\subsubsection{B.3.1 Details of the evaluation metrics.}
To comprehensively evaluate Y-Mol in drug design, we introduce \textbf{Valid}, \textbf{Unique}, \textbf{Novelty}, and \textbf{Diversity}. \textbf{Valid} assesses whether the generated drugs conform to the syntax rules of SMILES. Therefore, we adopt the RDKit~\cite{bento2020open} to parse the designed drugs, determining if the parsing process was successful. \textbf{Unique} calculates the proportion of non-repetitive drugs among the designed set of drugs, ensuring the model produces diverse molecules. \textbf{Novelty} signifies whether the generated drugs are previously unseen in the training dataset, preventing the model from re-designing existing drugs. \textbf{Diversity} describes the structural differences between designed drugs, calculated as:
\begin{equation}
    Diversity=1-\frac2{n(n-1)}\sum_{i=1}^n\sum_{j=1}^n sim(d_i,d_j),
\end{equation}
where $n$ is the number of designed drugs and $sim(\cdot,\cdot)$ indicates the Tanimoto distance based on the Morgan fingerprints. $d_i$ and $d_j$ represents the $i-$th and $j-$th drug within the set. 

\section{C Additional Experiments}
\subsection{C.1 Case of drug design on LLaMA2-7b}

We show some cases generated from LLaMA2-7b in Figure~\ref{fig:case_appendx}. As shown in Figure~\ref{fig:case_appendx}, we observe that Y-Mol performs well in both cases, while LLaMA2-7b can not understand the question well and answer some relevant context. This phenomenon indicates that Y-Mol learning large-scale biomedical knowledge can effectively enhance the capability of LLM in drug development.

\end{document}